\documentclass[runningheads]{llncs}
\usepackage{graphicx}
\usepackage{hhline}
\usepackage{color,soul}
\usepackage{float}
\usepackage{subcaption}
\usepackage{array}
\begin{document}
\newcolumntype{P}[1]{>{\centering\arraybackslash}p{#1}}
\title{Quantifying Uncertainty from Different Sources in Deep Neural Networks for Image Classification}
%
%
\author{A. Khoshsirat }
\authorrunning{}
%
\institute{Department of  Computer Science \& Engineering, Shiraz University, Shiraz, Iran 
}
\maketitle              
\begin{abstract}
Knowing about the confidence of an AI predictor is sometime essential to improve the safety and reliability of such system. Deep neural networks (NNs) are powerful predictors that have recently achieved very good performance on a wide spectrum of tasks. Quantifying predictive uncertainty in NNs is a challenging and yet on-going problem. Although there have been many efforts to equip NNs with tools to estimate uncertainty, many of the previous methods only focus on one of the three sources of uncertainty: Model uncertainty, Data uncertainty or Distributional uncertainty. In this paper we propose a complete framework to capture and detect all these three types of uncertainties for the task of image classification and demonstrate the efficiency of our method on popular image datasets.

\end{abstract}
\section{Introduction}

Deep neural networks (NNs) have achieved state-of-the-art performance on a wide variety of machine
learning tasks and are becoming increasingly popular in domains such as computer vision. DNNs have
proven to perform as good or even better than humans in certain scenarios and have come up with phenomenal results in areas like image analysis [1],
voice recognition [2] or even game strategies [3].
Due to this ongoing success, Deep Learning methods are being incorporated into
all kinds of applications. Research currently goes into using this
technology for self-driving cars [4] or for ID checks at passport controls [5].
However, when it comes to critical systems like these, we want to make sure that
our model does not make mistakes by any chance. We cannot expect
it to be always accurate or one-hundred percent perfect at their task, however, we
want it to let us know if it is not certain about a situation. This way, a second check
can be performed or the task can be passed to a human specialist. 
\par
Despite impressive accuracy in supervised learning benchmarks, NNs are poor at quantifying predictive
uncertainty, and tend to produce overconfident predictions. Classic Deep Learning networks do not incorporate uncertainty information but
only return a point prediction.
Classifiers failing to indicate when they are likely mistaken can limit their adoption or
cause serious accidents. For classification problems, we sometimes consider
a distribution over potential classes in order to express confidence for each class.
However, when faced with an unclassifiable input, i.e. an input that does not correspond to either class, results can be deceiving [6]. For example, a medical diagnosis model may consistently classify with
high confidence, even while it should flag difficult examples for human intervention.  Estimating uncertainty in a model’s predictions is important, as
it enables, for example, the safety of an AI system to be increased by acting on the model’s
prediction in an informed manner. This is crucial to applications where the cost of an error is high,
such as in autonomous vehicle control and medical, financial and legal fields.

Several research has been done in the field of uncertainty in machine learning. Different approaches have been published which enhance existing technologies by adding capabilities to networks which
enable them to not only give predictions or classification solutions but to also incorporate a
confidence measure. For modeling the uncertainty in these systems, one must consider different aspects of predictive uncertainty,
which results from three separate factors - model uncertainty, data uncertainty and distributional
uncertainty. Model uncertainty, or epistemic uncertainty , measures the uncertainty in estimating
the model parameters given the training data - this measures how well the model is matched to the
data. Data uncertainty, or
aleatoric uncertainty is uncertainty which arises from the natural complexity of the
data, such as class overlap and label noise.  Distributional uncertainty arises due
to mismatch between the training and test distributions (also called dataset shift). Taking all these three forms of uncertainties into account and being able to
distinguish between them for high risk applications, can help us choose what action to take to avoid fatal errors in the predictions of the system.

Although recent studies on deep neural networks have resulted in good detection of one form of uncertainty, none have fully anticipated all types of the known uncertainties. We've proposed a complete framework for deep neural networks to take these three kinds of uncertainties into account for the task of image classification. 

\section{Related Works}

Although there has been a lot of recent interest in adapting Deep Neural Networks to capture uncertainty, most of the methods focus on just one or two of the mentioned uncertainties and don't put any efforts into specification of the sources. One group of methods called Bayesian Neural Networks [7] concentrate on model uncertainty by specifying a prior
distribution upon the parameters of a NN and then, given the training data, the posterior
distribution over the parameters is computed. 
 
In a traditional generic neural networks we have fixed weights and biases that determine how an input is transformed into an output. In a Bayesian neural network, all weights and biases have a probability distribution attached to them. When the training is finished and the distributions are learned, the test data is given to the network more than one time. To classify an image, you do multiple runs (forward passes) of the network, each time with a new set of sampled weights and biases. Instead of a single set of output values what you get is multiple sets, one for each of the multiple runs. The set of output values represent a probability distribution on output values and hence you can find out confidence and uncertainty in each of the outputs. The variance of the distribution is then calculated for quantifying the uncertainty of the network for a specific test data. The current limitation of doing this work in large scale or real time production environments is posterior computation. Variational inference techniques and/or efficient sampling methods to obtain posterior are computational demanding. This problem has lead the researchers to work on finding other methods with less computational and time cost [8].

One recent development on estimating uncertainty, has been the technique of Monte-Carlo Dropout [9]. Dropout is usually applied for deep neural networks in order to
avoid over-fitting. Also originally, dropout is only used during training time and
dismissed at prediction time. Monte-Carlo Dropout , however, uses this method
to introduce randomness to the prediction process. This way, the network can be
evaluated for the same input multiple times, resulting in a set of predictions that can
be used to estimate a distribution in the target domain or to determine statistical
values. 
At prediction time, this technique estimates predictive uncertainty using an ensemble of multiple stochastic forward passes and
computing the mean and spread of the ensemble. The mean prediction can
be estimated by averaging the forward passes while the uncertainty can be estimated
in terms of the empirical variance. This method has been successfully applied to
tasks in computer vision [10].

Another method that has been proposed recently, uses an ensemble of deep neural networks for uncertainty estimation. Deep Ensembles [11] take a different approach than previous methods: it assumes the data to have a given parametrized distribution where the parameters
depend on the input. Finding these parameters is the aim of the training process,
i.e. the prediction network will not output a single value but instead will output
the distributional parameters for the given input. In other words, instead of training one network, an ensemble of identical networks are trained with parameter sharing and data set splitting. At prediction time, the individual distributions are then
averaged resulting in the final estimate. The outputs distribution (its variance) can then be used for estimating the uncertainty of the system. This method has been tested on regression problems.

Although these methods have resulted in good estimates for predictive uncertainty, they can not specify the source of it. For instance, in a classification problem, a test data with high variance of outputs distribution in a Monte-Carlo Dropout network, can either be interpreted as, an intrinsically hard to classify sample of the known data distribution, that is near a decision boundary (data uncertainty), or an out-of-distribution sample which has not been introduced to the network in training time (distributional uncertainty).

On the other end of the spectrum, some methods have been proposed to specifically estimate one of the three kinds of uncertainty for a test data point in DNNs. A recent study, uses deep autoencoders for detecting out-of-distribution samples at test time on image data [12]. Autoencoders optimize the compression of
input data to a latent space of a smaller dimension and
attempt to accurately reconstruct the original input using the features that has been learned in the latent space.
Since the latent vector is optimized to capture the salient features from the inlier
class only, it is assumed that images of objects from outside of the
training classes cannot effectively be compressed and reconstructed. The study proposes using deep autoencoders and then a clustering routine to detect OOD samples at test time. Although it hasn't directly stated, a form of distributional uncertainty is used for this detection which has been produced by computing the reconstruction error. 

In another study, misclassified examples detection is done by using the softmax probabilities for classification deep neural networks [13].  Correctly classified examples tend to have greater maximum softmax probabilities than erroneously classified examples, allowing for their detection. This also can be considered a use of data uncertainty, since the softmax layer focuses on the inlier classes and their boundaries with each other.
\\ \\

\section{Proposed Method}

As mentioned in the previous section, different methods have been proposed to capture different types of uncertainty in DNNs. In this paper we propose a simple and functional framework which combines the previously proposed methods and also new methodologies to better capture the different predictive uncertainties for the task of image classification. This framework includes an ensemble of  DNNs for model uncertainty, a supervised reconstruction auto-encoder for distributional uncertainty and using the softplus activation function in the last layer of the DNN for capturing the data uncertainties, working side by side to model the three types of uncertainties. Finally our framework is evaluated on detection of misclassified or out-of-distribution samples for classification of various image datasets. \\ \\
In the sections below we describe our proposed methods for any of the three types of the mentioned uncertainties.

\subsection{Model Uncertainty}
Model or epistemic uncertainty is about how well a used model is matched to our training data for the problem at hand. Although Bayesian methods can model this uncertainty for NNs, Their computational cost and complexity can somehow affect the system performance, therefore using non-Bayesian methods are encouraged. 

Here we use an ensemble of deep neural networks with the same architecture (suitable for the classification task) and random initialization to present the parameter uncertainties of the model.
This method can show how uncertain we are for a deep neural network model on a specific data set. 

\begin{figure}[H]
\centering
\includegraphics[width=0.85\textwidth]{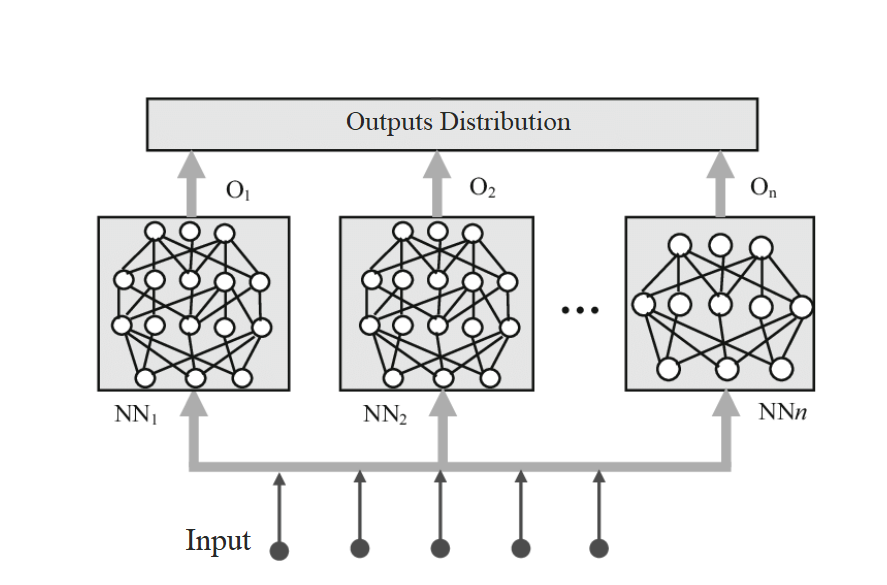}
\caption{Ensemble of DNNs, the proposed architecture for estimating model uncertainty.}
\end{figure}
Due to different initialization of the networks' parameters, an unsuitable network could result in more dissimilar output results in the ensemble. Hence the spread of the distribution on the outputs, can be used to estimate how well the model is fitting for the classification task.  This method is simple and parallelizable compared to Bayesian methods and can very well be applied to different types of NNs for this task. In this scenario, we've used ten Convolutional Neural Networks (CNN) having the same architecture, but different initialization (random) for image classification.

\subsection{Data Uncertainty}
Data uncertainty or aleatoric uncertainty is the complexity of the data itself. Modeling and detecting this type of uncertainty can help us predict whether a test sample is classified correctly and how confident we are for any given prediction. 

As mentioned in the previous section, some works have been done to use softmax probabilities for somehow capturing the data uncertainty specifically and detecting the probable misclassified samples. However, softmax activation func. limits the probabilities of the last layer of the network to always sum up to 1. Therefore, during training, the network is bound to decrease the other classes' probabilities in order to increase the maximum probable class for a specific training data. This limitation can lead to missing the beneficial information about the data for capturing the accurate data uncertainty. 

To address this issue, instead of softmax, we use the Softplus activation function in the last layer of a DNN. The softplus func. has no such limitations on the last layer's neurons and can better show the data points probability information.
For computing the data uncertainty metric at test time, We take the softplus probabilities and compute two terms. First we subtract the maximum prob. by the second maximum prob. for a test data point. This value is larger for the samples with less data uncertainty; confident network predictions. Secondly, we sum up all the other classes' probabilities. This value shows the spread of the class probabilities which can be interpreted as the uncertainty of the networks decision to classify a test data point to a specific class. Finally we divide these two terms to compute a suitable metric for networks data uncertainty for predictions.
Using proper thresholds, we can very well detect the hard to classify or probably will-be-misclassified samples.
The final formula for computing data uncertainty is shown below:
\begin{equation}
Data \; uncertainty = \frac{\sum_{i=3}^{n} p_{sp}(c_{i})}{p_{sp}(c_{1})-p_{sp}(c_{2})}
\end{equation}
In this equation, $p_{sp}$ is the probability of softplus layer and the class probabilities are sorted from the maximum to minimum ($c_{1}$,$c_{2}$,...,$c_{n}$) for a specific data point. 

\subsection{Distributional Uncertainty}
This uncertainty arises when the samples given at test time are not from the distribution of the training data. For detecting this, we propose using a supervised reconstruction autoencoder. Reconstruction autoencoders are NNs that compress the data to a bottleneck layer (encoder) and then reconstruct the input data from that layer (decoder). This would force the NN to learn the distribution of the input data in the bottleneck layer. As previously mentioned, this tool can be used to detect out of distribution samples at test time.
\begin{figure}[H]
\centering
\includegraphics[width=0.85\textwidth]{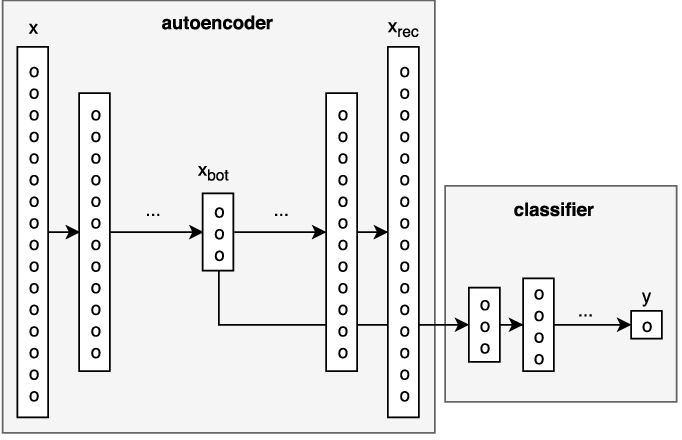}
\caption{Architecture used for supervised reconstruction autoencoder. This network is trained by minimizing both the classification and reconstruction loss of the labeled input data.}
\end{figure}

Using the supervised version of autoencoder can guarantee the distinction of the class distributions in the bottleneck layer as well and therefore can result in better detection of OOD samples from unknown classes at test time. OOD sample detection is done by computing the reconstruction loss for each test sample at test time, and comparing the loss to a threshold. Figure 1 shows the architecture used for the supervised autoencoder.
Another method that we used to capture this uncertainty is equipping the NNs in the ensemble with an 'unknown' class and then adding various data from other distributions with 'unknown' label to the training data. Training each NN in the ensemble with different 'unknown' labeled data gives the best result for detecting OOD samples.
\\ \\
Figure 2 shows the complete diagram of the proposed framework. Using proper thresholds, the framework can be used for detecting misclassified or OOD samples for image classification.

\begin{figure}[H]
\centering
\includegraphics[width=0.99\textwidth]{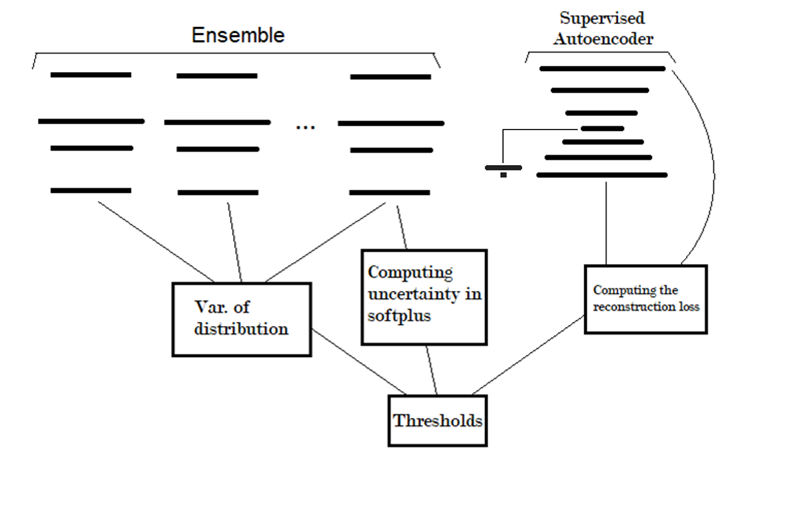}
\caption{The diagram of the complete proposed framework}
\end{figure}

\section{Experiments}
The proposed framework can capture all three kinds of uncertainty. For each one, we've implemented various tests to show the efficiency and higher accuracy of the proposed method compared to the previous works. 
\\ \\ 
First the results of data uncertainty are presented. In figures 4 and 5, we can see the test data points of two datasets (Cifar10 and MNIST) that have large data uncertainty, detected with the proposed method. As one can see, these instances are hard to classify for the model because they look like they can originate from two or more classes. For example, we've detected the planes that are much like birds, which are two classes of data in Cifar10 data set. 

\begin{figure}[H]
\centering
\includegraphics[width=0.85\textwidth]{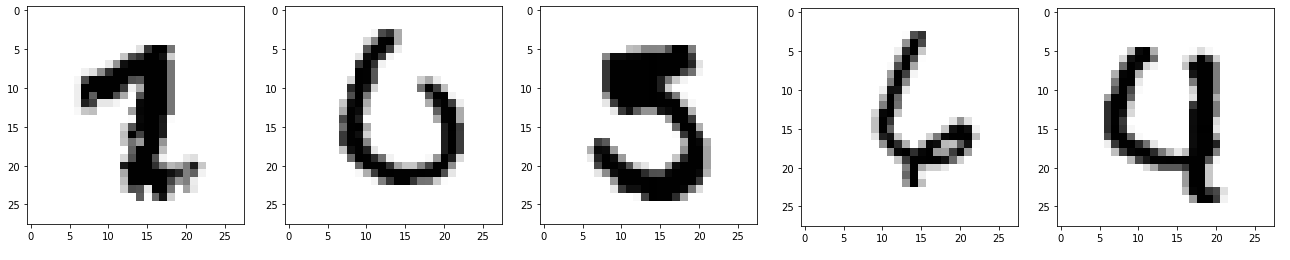}
\caption{Test data with high data uncertainty for MNIST dataset.}
\end{figure}
\begin{figure}[H]
\centering
\includegraphics[width=0.85\textwidth]{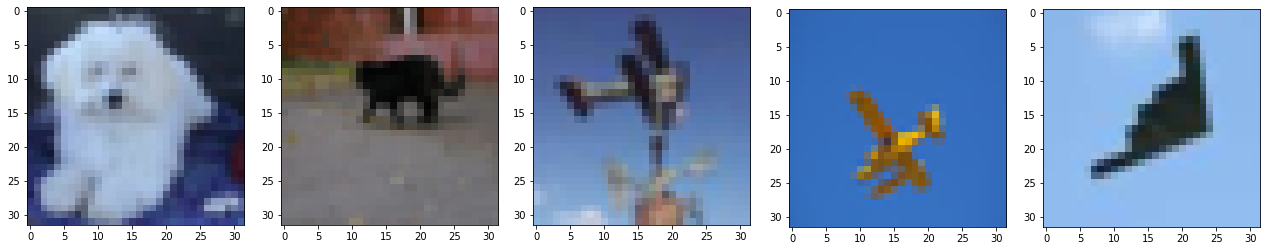}
\caption{Test data with high data uncertainty for Cifar10 dataset.}
\end{figure}

A common use of data uncertainty is the detection of misclassified test samples beforehand. For this test, we've compared the proposed method to MC-Dropout on the ability to detect the hard to classify samples. Figure 6 shows the T-SNE  results of the second last layer of a classification network trained on MNIST. As we can see, visualizing this layer can very well demonstrate the different classes' distributions and clusters of the test data. In figures 8, we see the results of misclassified samples detection for the test data using the proposed method and MC-Dropout. As we can see, the proposed method tends to be more accurate in finding the between class or wrongly placed test samples. Moreover, MC-Dropout have more false positives compared to the proposed methods for some classes. This can be confirmed in the numeric results later in this section.

\begin{figure}[H]
\centering
\includegraphics[width=0.75\textwidth]{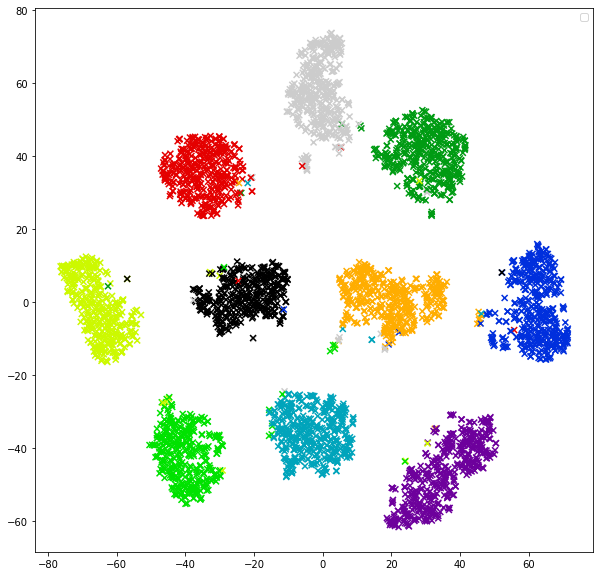}
\caption{The T-SNE visualization of the second last layer in a network trained on MNIST data set, for test samples.}
\end{figure}
\begin{figure}[H]
\centering
\begin{subfigure}[b]{0.75\textwidth}
   \includegraphics[width=1\linewidth]{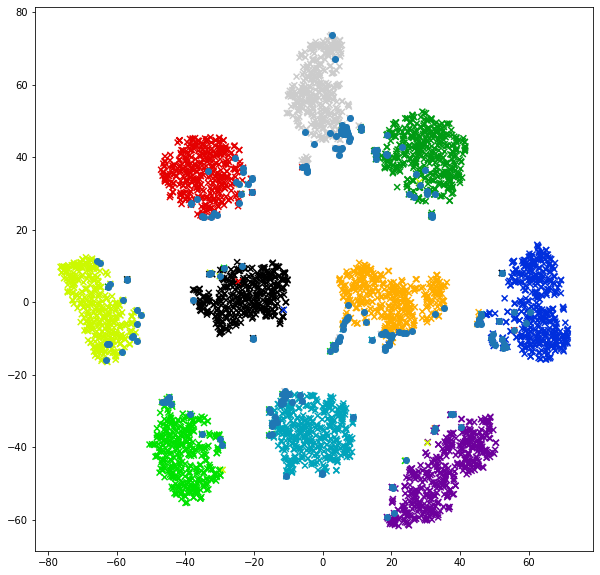}
   \caption{}
   \label{fig:Ng1} 
\end{subfigure}

\begin{subfigure}[b]{0.75\textwidth}
   \includegraphics[width=1\linewidth]{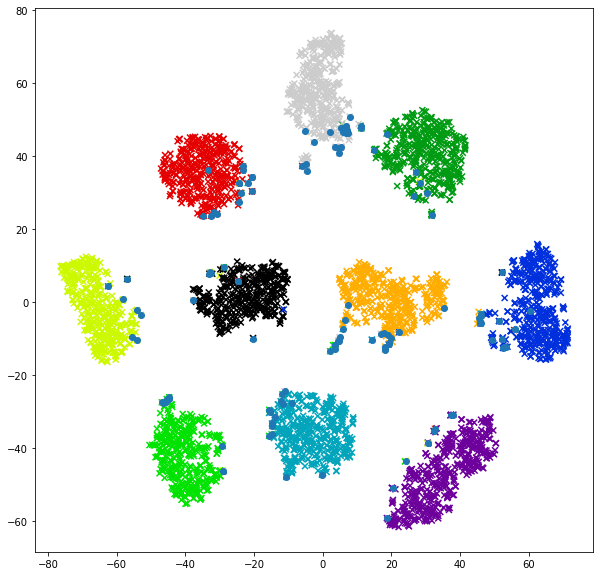}
   \caption{}
   \label{fig:Ng2}
\end{subfigure}
\caption{The T-SNE visualization of the second last layer in a network trained on MNIST data set, for test samples. The detected misclassified samples with MC-Droput method (a) and the proposed method (b) are painted blue.}
\end{figure}

 For Cifar10 data set, Figure 9 shows the T-SNE  results of the second last layer. We've also showed the results of misclassified samples detection for the test data using the proposed method and MC-Dropout on Cifar10 in figure 10. In this task, the proposed method does a lot better in finding the between class or wrongly placed test samples than MC-Dropout, with less false positives and more true positives. The MC-Dropout method seems to act more strict and have more false positives in complex datasets.

\begin{figure}[H]
\centering
\includegraphics[width=0.90\textwidth]{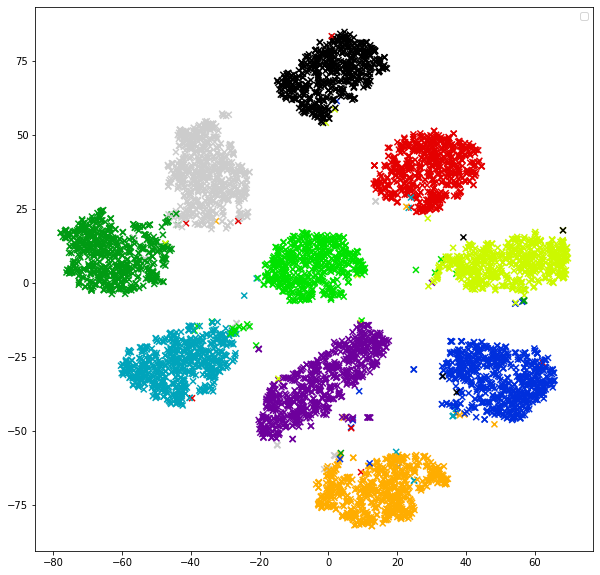}
\caption{The T-SNE visualization of the second last layer in a network trained on Cifar10 data set, for test samples.}
\end{figure}
\begin{figure}[H]
\centering
\begin{subfigure}[b]{0.75\textwidth}
   \includegraphics[width=1\linewidth]{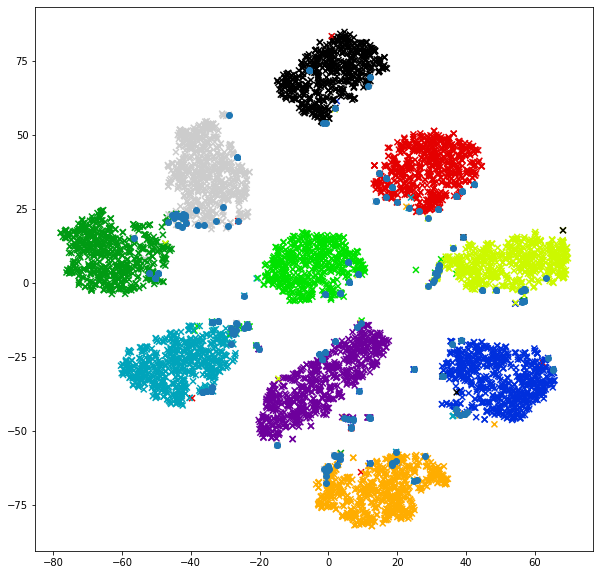}
   \caption{}
   \label{fig:Ng1} 
\end{subfigure}

\begin{subfigure}[b]{0.75\textwidth}
   \includegraphics[width=1\linewidth]{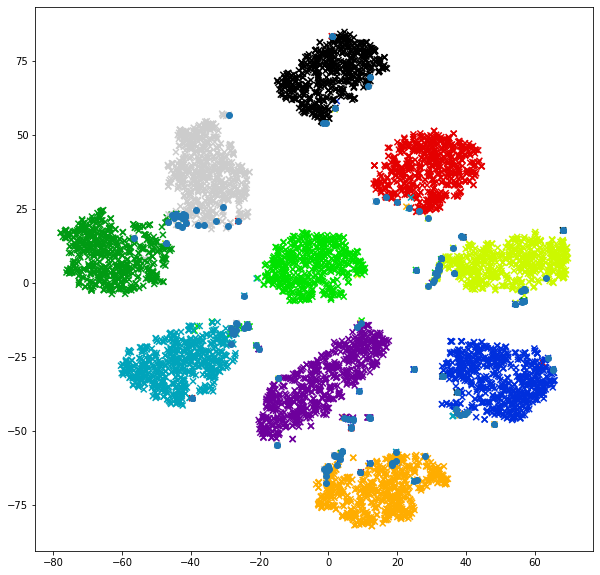}
   \caption{}
   \label{fig:Ng2}
\end{subfigure}
\caption{The T-SNE visualization of the second last layer in a network trained on Cifar10 data set, for test samples. The detected misclassified samples with MC-Droput method (a) and the proposed method (b) are painted blue.}
\end{figure}

For misclassified detection test, we've compared our proposed method to MC-Dropout and also the baseline method which uses softmax to compute the data uncertainty on various image classification datasets. The datasets we've used include: Mnist ,Fashion-Mnist [17], Cifar10, Cifar100 [18] and TinyImageNet [19]. The metrics used in this paper are True Positive Rate and False Positive Rate [20]. These metrics are commonly used for tasks that include a threshold for detection which is the case in misclassified samples detection.

\begin{center}
\begin{tabular}{ |P{3.6cm}||P{2.4cm}|P{2.4cm}|   }
 \hline
 \multicolumn{3}{|c|}{Mnist} \\
 \hline
   & TPR & FPR \\
 \hline
 Baseline (Softmax)  &  77.6 \%     & 2.9 \% \\
 MC-Dropout          &  80.2 \%     & 7.4 \%  \\
 Proposed            &  \textbf{84.1} \%     & \textbf{2.3} \% \\
 \hline
\end{tabular}

\hfill \break

\begin{tabular}{ |P{3.6cm}||P{2.4cm}|P{2.4cm}|   }
 \hline
 \multicolumn{3}{|c|}{Fashion-Mnist} \\
 \hline
   & TPR & FPR \\
 \hline
 Baseline (Softmax)  &  90.32 \%    & \textbf{3.6} \% \\
 MC-Dropout          &  88.3 \%     & 4.3 \%  \\
 Proposed            &  \textbf{92.61} \%    & 3.8 \% \\

 \hline
\end{tabular}

\hfill \break

\begin{tabular}{ |P{3.6cm}||P{2.4cm}|P{2.4cm}|   }
 \hline
 \multicolumn{3}{|c|}{Cifar10} \\
 \hline
   & TPR & FPR \\
 \hline
 Baseline (Softmax)  &  85.8 \%     & 9.5 \% \\
 MC-Dropout          &  82.1 \%     & 10.1 \%  \\
 Proposed            &  \textbf{87.6} \%     & \textbf{8.2} \% \\

 \hline
\end{tabular}

\hfill \break

\begin{tabular}{ |P{3.6cm}||P{2.4cm}|P{2.4cm}|   }
 \hline
 \multicolumn{3}{|c|}{Cifar100} \\
 \hline
   & TPR & FPR \\
 \hline
 Baseline (Softmax)  &  55.68 \%     & \textbf{26.0} \% \\
 MC-Dropout          &  60.54 \%     & 32.7 \%  \\
 Proposed            &  \textbf{64.21} \%     & 29.5 \% \\

 \hline
\end{tabular}

\hfill \break

\begin{tabular}{ |P{3.6cm}||P{2.4cm}|P{2.4cm}|   }
 \hline
 \multicolumn{3}{|c|}{TinyImageNet} \\
 \hline
   & TPR & FPR \\
 \hline
 Baseline (Softmax)  &  53.81 \%    & 42.9 \% \\
 MC-Dropout          &  58.74 \%     & \textbf{28.1} \%  \\
 Proposed            &  \textbf{59.01} \%     & 30.4 \% \\

 \hline
\end{tabular}
\end{center}

For comparing the ability to capture distributional uncertainty, we have tested our framework on the datasets mentioned above for OOD samples detection.  The OMNIGLOT dataset [21], scaled down to 28x28 pixels, was used as real ’OOD’ data for MNIST. For OOD of Cifar10, we've used TinyImageNet data set which is identical to Cifar10. As shown in Figure 3, the autoencoder can reconstruct the numbers found in the training data much more accurately than the OOD samples. These examples shows the efficiency of the model in OOD detection.

\begin{figure}[H]
\centering
\begin{subfigure}[b]{0.45\textwidth}
   \includegraphics[width=1\linewidth]{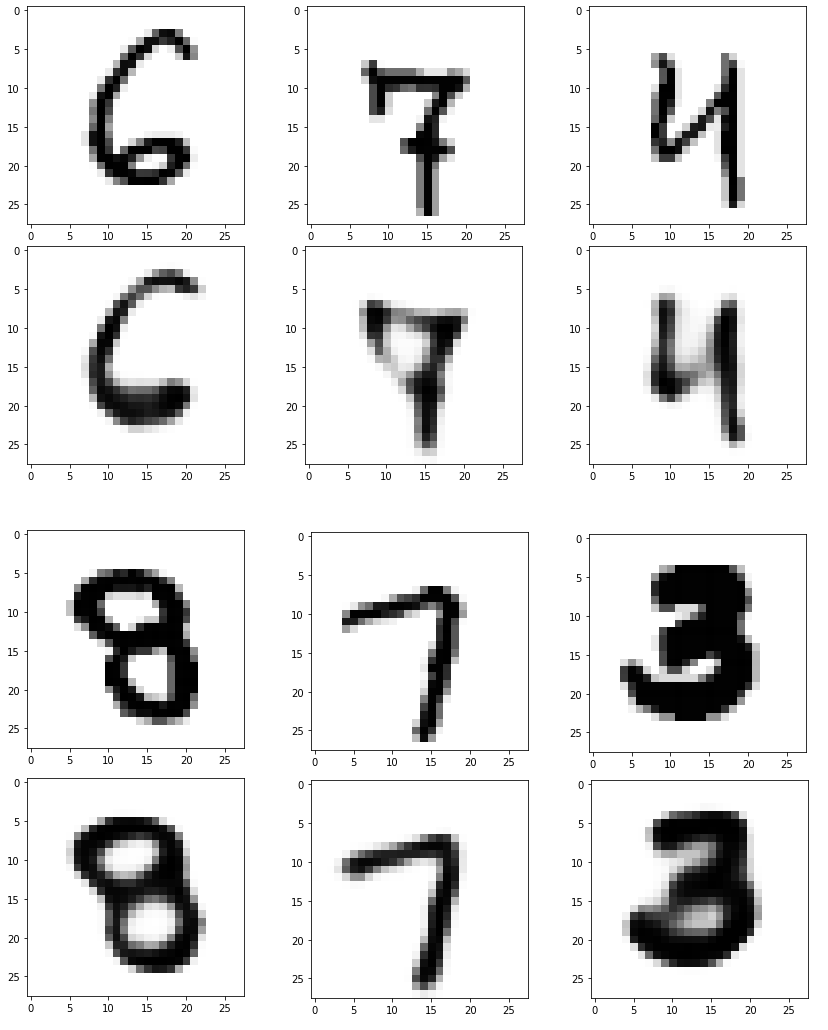}
   \caption{}
   \label{fig:Ng1} 
\end{subfigure}

\begin{subfigure}[b]{0.45\textwidth}
   \includegraphics[width=1\linewidth]{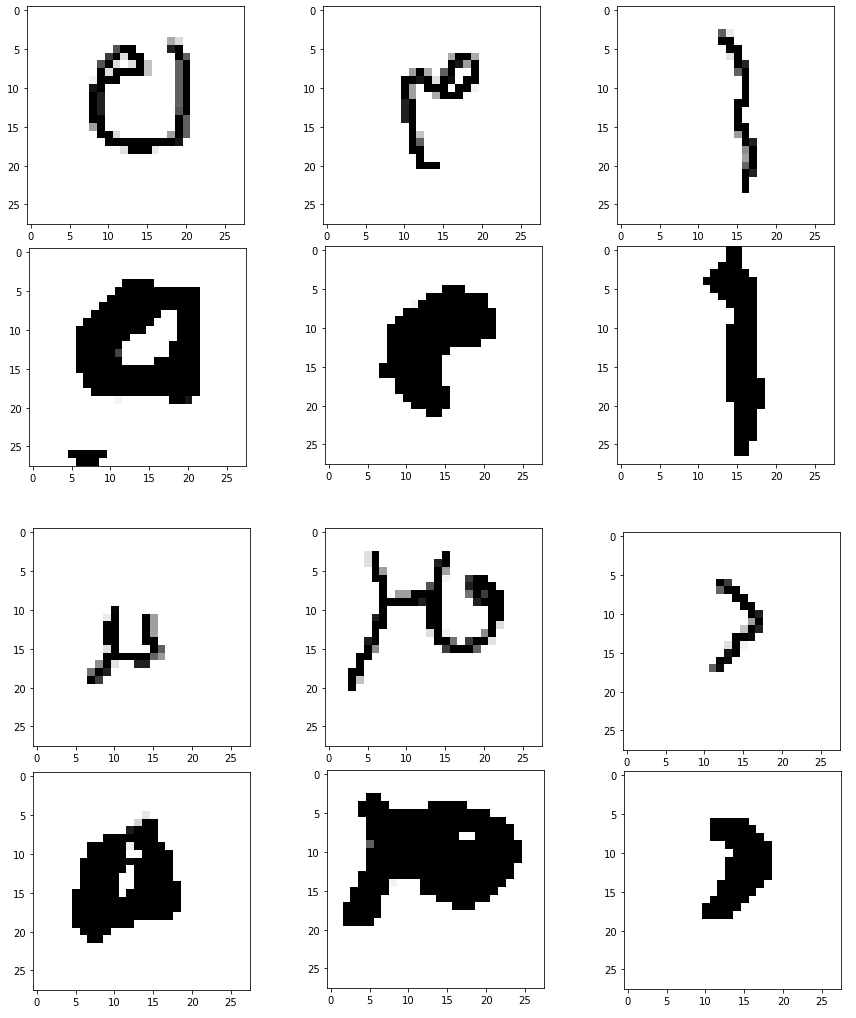}
   \caption{}
   \label{fig:Ng2}
\end{subfigure}
\caption{The input and output of the supervised reconstruction autoencoder (trained on MNIST) for some MNIST (a) and OMNIGLOT (b) samples. The OMNIGLOT (OOD) samples are reconstructed poorly, indicating they're from a different distribution than training data.}
\end{figure}

\begin{figure}[H]
\centering
\begin{subfigure}[b]{0.60\textwidth}
   \includegraphics[width=1\linewidth]{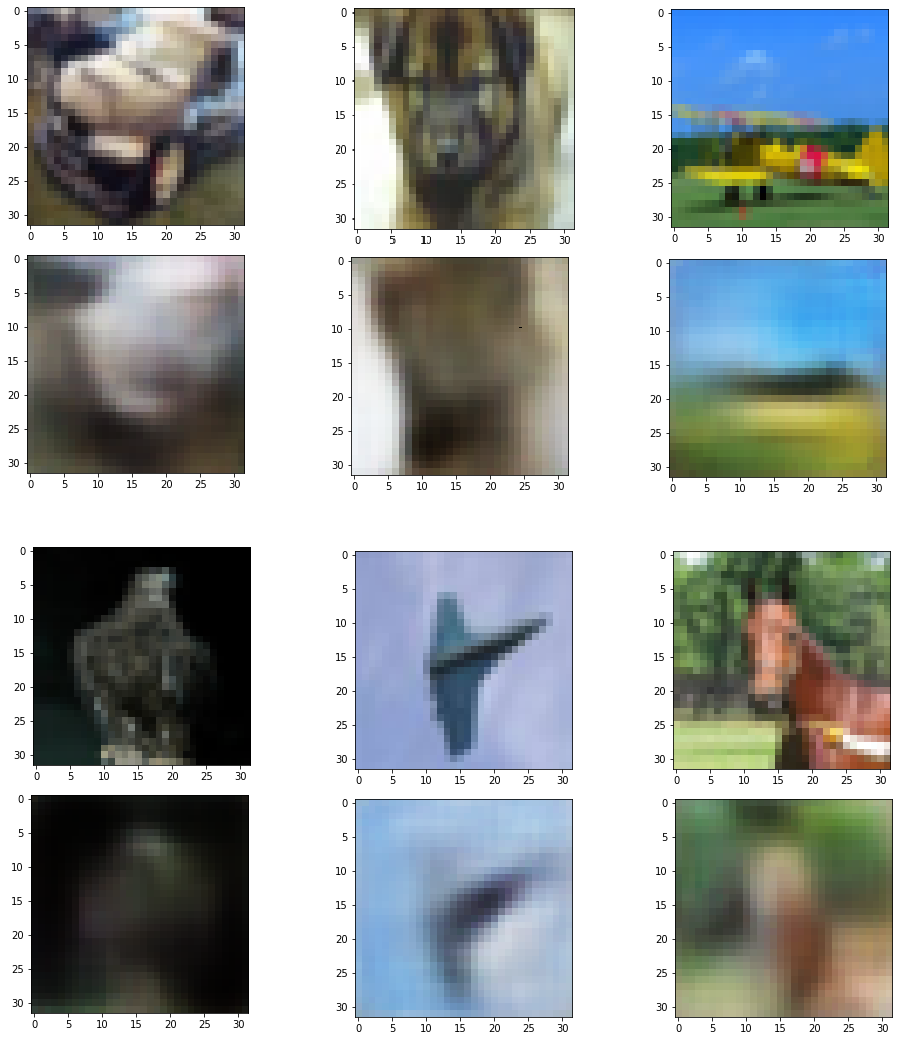}
   \caption{}
   \label{fig:Ng1} 
\end{subfigure}

\begin{subfigure}[b]{0.60\textwidth}
   \includegraphics[width=1\linewidth]{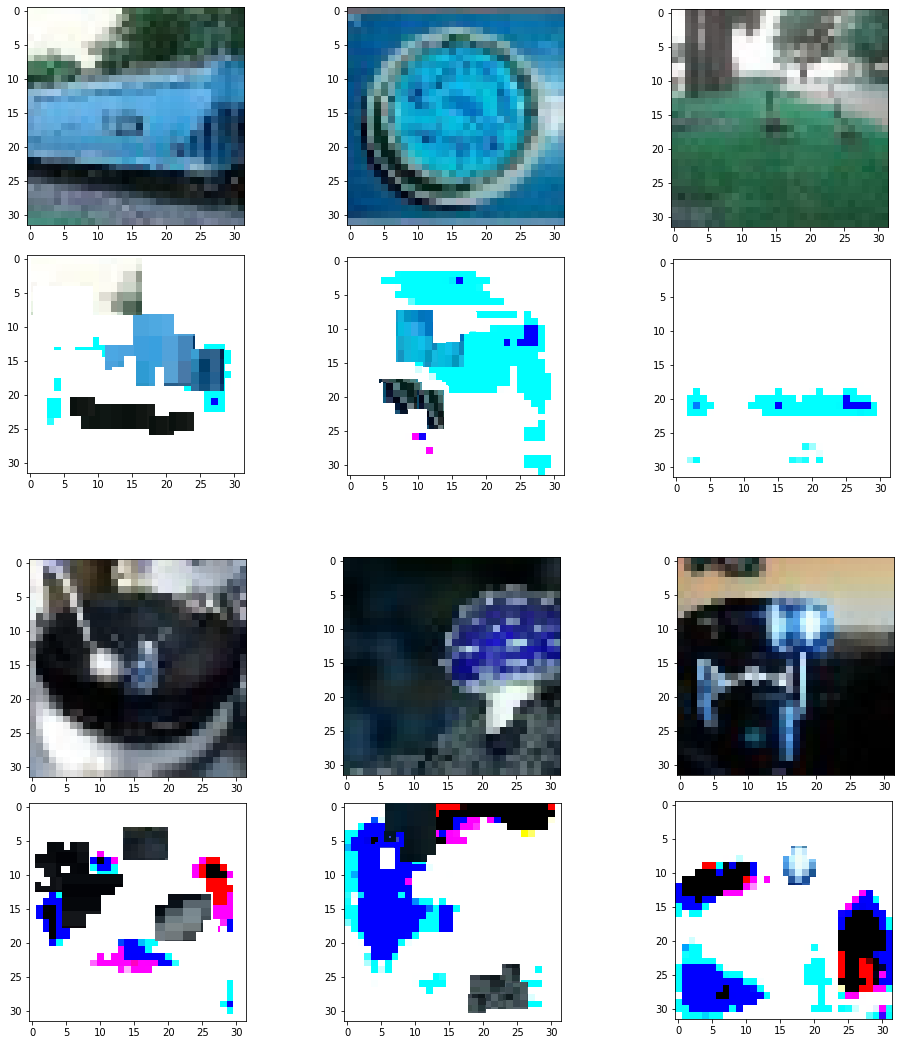}
   \caption{}
   \label{fig:Ng2}
\end{subfigure}
\caption{The input and output of the supervised reconstruction autoencoder (trained on Cifar10) for a few Cifar10 (a) and TinyImageNet (b) samples. The TinyImageNet (OOD) samples are reconstructed poorly, indicating they're from a different distribution than training data.}
\end{figure}

\begin{tabular}{ |P{3.8cm}||P{2cm}|P{2cm}|P{2cm}|   }
 \hline
 \multicolumn{4}{|c|}{Mnist-Omniglot} \\
 \hline
   & TPR & FPR & False-Neg\\
 \hline
 Baseline (Autoencoder)  & 96.6 \%    & 5.25 \% &   188\\
 MC-Dropout&   98.9 \%  & 4.12 \%   & 154\\
 Proposed & \textbf{99.2} \% & \textbf{4.01} \%&  132 \\

 \hline
\end{tabular}

\begin{tabular}{ |P{3.8cm}||P{2cm}|P{2cm}|P{2cm}|  }
 \hline
 \multicolumn{4}{|c|}{Fashion-Mnist-Omniglot} \\
 \hline
   & TPR & FPR & False-Neg\\
 \hline
 Baseline (Autoencoder)  & 84.1 \%    & \textbf{7.21} \% &   234\\
 MC-Dropout&             87.3 \%  & 9.01 \%   & 198\\
 Proposed &             \textbf{92.2} \% & 9.62 \%&  173 \\

 \hline
\end{tabular}

\begin{tabular}{ |P{3.8cm}||P{2cm}|P{2cm}|P{2cm}|  }
 \hline
 \multicolumn{4}{|c|}{Cifar10-TinyImageNet} \\
 \hline
   & TPR & FPR & False-Neg\\
 \hline
 Baseline (Autoencoder)  & 91.5 \%    & 17.01 \% &   357\\
 MC-Dropout&   90.2 \%  & \textbf{15.12} \%   & 401\\
 Proposed & \textbf{95.0} \% & 15.40 \%&  273 \\

 \hline
\end{tabular}

\begin{tabular}{  |P{3.8cm}||P{2cm}|P{2cm}|P{2cm}|}
 \hline
 \multicolumn{4}{|c|}{Cifar100-TinyImageNet} \\
 \hline
   & TPR & FPR & False-Neg\\
 \hline
 Baseline (Autoencoder)  & 56.5 \%    & 29.0 \% &   427\\
 MC-Dropout&   63.7 \%  & 37.4 \%   & 304\\
 Proposed & \textbf{66.8} \% & \textbf{26.2} \%&  282 \\

 \hline
\end{tabular}

\begin{tabular}{  |P{3.8cm}||P{2cm}|P{2cm}|P{2cm}|}
 \hline
 \multicolumn{4}{|c|}{Cifar50-Cifar50} \\
 \hline
   & TPR & FPR & False-Neg\\
 \hline
 Baseline (Autoencoder)  & 57.3 \%    & 39.4 \% &   562\\
 MC-Dropout&   \textbf{64.2} \%  & 41.9 \%   & 445\\
 Proposed & 63.7 \% & \textbf{37.6} \%&  483 \\

 \hline
\end{tabular}

Finally we've tested our proposed method in case of computing the model uncertainty. For this experiment, we've created different scenarios of changing a model for a classification problem to observe the behaviour of the computed model uncertainty. For each specific model, we've implemented an ensemble of the same networks to capture the distribution on the outputs and interpret the spread of the distribution as model uncertainty.
\\ \\
At first, we've changed the number of layers of a CNN for MNIST classification in each step and then computed the distribution on the ensemble's outputs. In figure 11, we can see that for larger number of layers the distribution is sharper, so the model uncertainty decreases, which can be interpreted as if the model has better capacity for the classification problem at hand. 

\begin{figure}[H]
\centering
\includegraphics[width=0.75\textwidth]{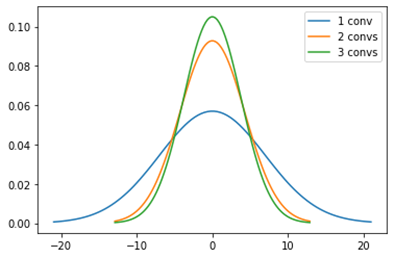}
\caption{Distribution of test data outputs on MNIST for different number of layers. Note that the variance of the distribution is the model uncertainty for every scenario}
\end{figure}

In the next experiment, we've changed the network's loss function  in each step. In figure 12, we can see the different model uncertainties (variance of the distribution) for models trained with different loss functions. The two loss functions specified for a multi-class classification problem gives us the least model uncertainty, which is logically accurate. The one with the least model uncertainty, categorical cross-entropy function, is commonly used for classification of MNIST [16].
\begin{figure}[H]
\centering
\includegraphics[width=0.75\textwidth]{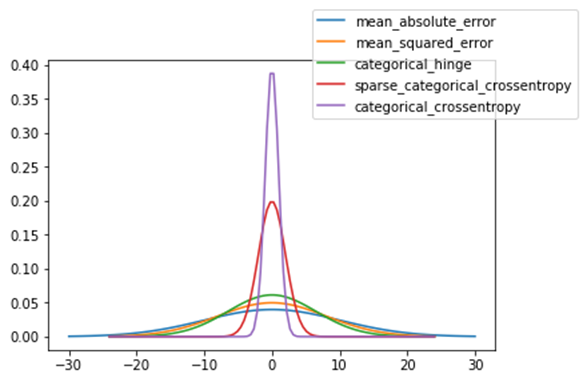}
\caption{Distribution of test data outputs on MNIST for different loss functions. Note that the variance of the distributions are the model uncertainty for every loss function used.}
\end{figure}
Next, we've tested various number of hidden layer neurons of the networks. The model uncertainty seems to be decreasing rapidly, up until the number of neurons reach the satisfying number of 16 for classification of MNIST data set. This also shows the accuracy of the computed model uncertainties.

\begin{figure}[H]
\centering
\includegraphics[width=0.75\textwidth]{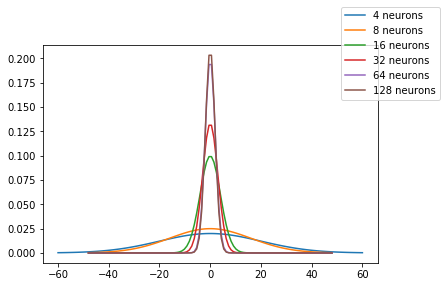}
\caption{Distribution of test data outputs on MNIST for different Number of hidden layers' neurons.The variance of the distributions are the model uncertainty for every number of hidden neurons used.}
\end{figure}
 
 Finally, we experimented different number of training epochs. For a few epochs, the model tend to have high uncertainty due to the fact that it hasn't learned much from the data and the random first initialization are not changed much. For 4 epochs, which is enough for networks' training, and higher numbers the model uncertainty decreases to a minimum.

\begin{figure}[H]
\centering
\includegraphics[width=0.75\textwidth]{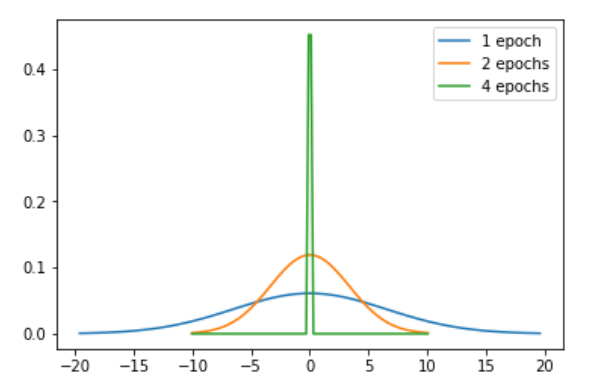}
\caption{Distribution of test data outputs on MNIST for different Number of training epochs.The variance of the distributions are the model uncertainty for every scenario.}
\end{figure}

\subsection{Conclusion}
This work describes the limitations of previous works on uncertainty estimations as a matter of sources of uncertainty and proposes to treat the three known types of uncertainty in a complete framework. We have presented a novel
method,  which allows data, distributional and model uncertainty to be treated separately within a consistent interpretable framework. The proposed method is shown to yield
more accurate estimates of data uncertainty and distributional uncertainty than MC Dropout and standard baseline softmax probabilities and autoencoders on the task
of OOD and misclassified samples detection for various image datasets. Uncertainty measures can be analytically calculated at test time in one pass in the proposed framework, reducing computational cost relative to previous Bayesian or dropout approaches. Having investigated the framework for image classification, it is interesting to apply them to other tasks in computer vision.

%
%
%
%

\end{document}